\documentclass{article} %

\makeatletter
\def\author@and{\\[\affilsep]\setcounter{Maxaffil}{0}} %
\makeatother

\usepackage{fullpage,times}

\usepackage{amsmath,amsfonts,bm}

\def\eqref#1{equation~\ref{#1}}

\def\1{\bm{1}}

\DeclareMathAlphabet{\mathsfit}{\encodingdefault}{\sfdefault}{m}{sl}
\SetMathAlphabet{\mathsfit}{bold}{\encodingdefault}{\sfdefault}{bx}{n}

\usepackage{lmodern}
\usepackage{xspace}
\usepackage[protrusion=true,expansion=true]{microtype}
\usepackage{amsfonts,amsmath,amssymb,amsthm, pbox}

\usepackage[colorlinks,citecolor=blue,bookmarks=true]{hyperref}

\usepackage{multirow}
\usepackage{dsfont} %
\usepackage{algorithmic,algorithm}

\usepackage[shortlabels]{enumitem}
\setitemize{noitemsep,topsep=0pt,parsep=0pt,partopsep=0pt}
\setenumerate{noitemsep,topsep=0pt,parsep=0pt,partopsep=0pt}

\makeatletter
\@ifundefined{theorem}{%
  \theoremstyle{definition}
  
  \theoremstyle{plain}

  \theoremstyle{remark}

}{}
\makeatother

\newcommand{\ignore}[1]{}

\newcommand{\EE}{\mathbb{E}}

\def \Paren#1{{\left({#1}\right)}}

\def\ignore#1{}

\newcommand{\bi}{\begin{itemize}}
\newcommand{\ei}{\end{itemize}}

\def\orpro{\mathop{\mathchoice
   {\vee\kern-.49em\raise.7ex\hbox{$\cdot$}\kern.4em}
   {\vee\kern-.45em\raise.63ex\hbox{$\cdot$}\kern.2em}
   {\vee\kern-.4em\raise.3ex\hbox{$\cdot$}\kern.1em}
   {\vee\kern-.35em\raise2.2ex\hbox{$\cdot$}\kern.1em}}\limits}

\def\andpro{\mathop{\mathchoice
 {\wedge\kern-.46em\lower.69ex\hbox{$\cdot$}\kern.3em}
 {\wedge\kern-.46em\lower.58ex\hbox{$\cdot$}\kern.25em}
 {\wedge\kern-.38em\lower.5ex\hbox{$\cdot$}\kern.1em}
 {\wedge\kern-.3em\lower.5ex\hbox{$\cdot$}\kern.1em}}\limits}

\def\simge{\mathrel{%
   \rlap{\raise 0.511ex \hbox{$>$}}{\lower 0.511ex \hbox{$\sim$}}}}

\def\simle{\mathrel{
   \rlap{\raise 0.511ex \hbox{$<$}}{\lower 0.511ex \hbox{$\sim$}}}}

\newcommand{\tx}{\ensuremath{M}}

\usepackage{booktabs}
\usepackage{natbib}
\usepackage{graphicx}
\usepackage{pdfpages}
\usepackage{numprint}
\usepackage{tikz}
\usepackage{subcaption}
\usepackage{siunitx}
\usepackage[table]{xcolor}
\usepackage{graphicx}
\usetikzlibrary{positioning, shapes, arrows}
\usepackage{amssymb}   %
\usepackage{makecell}  %
\usepackage{paralist} %

\def\withnotes{1}
\def\withunfinished{1}
\def\withcolor{1}

\ifnum\withcolor=1

\else

\fi

\ifnum\withnotes=1
\newcommand{\zs}[1]{{\noindent \textit{\small\textcolor{red}{ziteng: #1}}}}
\newcommand{\ts}[1]{{\noindent \textit{\small\textcolor{red}{theertha: #1}}}}
\newcommand{\adrian}[1]{{\noindent \textit{\small\textcolor{red}{adrian: #1}}}}
\newcommand{\vikas}[1]{{\noindent \textit{\small\textcolor{blue}{vikas: #1}}}}
\newcommand{\todo}[1]{{\noindent \textit{\small\textcolor{blue}{TODO: #1}}}}

\else
\newcommand{\zs}[1]{}
\newcommand{\ts}[1]{}
\newcommand{\adrian}[1]{}
\newcommand{\vikas}[1]{}
\newcommand{\todo}[1]{}
\fi

\ifnum\withunfinished=1
\newcommand{\unfinished}[1]{{#1}}
\else
\newcommand{\unfinished}[1]{{}}
\fi

\renewcommand{\ignore}[1]{}

\newcommand{\cafeq}{{\rm CafeQ}}

\newcommand{\uniform}{{\textsc{uniform}}}
\newcommand{\random}{{\textsc{random}}}
\newcommand{\pqe}{PQE\xspace}
\usepackage[capitalise]{cleveref}
\crefname{claim}{Claim}{Claim}
\newcommand{\logsumexp}{{\rm LogSumExp}}
\usepackage{hyperref}
\usepackage{url}
\usepackage{sidecap}
\usepackage{authblk}

\title{CafeQ: Calibration-free Quantization via Learned Transformations and Adaptive Rounding}

\author{Ziteng Sun$^\dag$, Adrian Benton$^\dag$, Samuel Kushnir$^\dag$, Asher Trockman$^\dag$ \\ Vikas Singh$^\circ$, Suhas Diggavi$^\ddag$, Ananda Theertha Suresh$^\dag$ \\
\medskip
  $\dag$ Google: \texttt{\{zitengsun,adbenton,samuelkushnir,trockman,theertha\}@google.com} \\
  $\circ$ University of Wisconsin, Madison: \texttt{vsingh@biostat.wisc.edu} \\
  $\ddag$ University of California, Los Angeles: \texttt{suhas@ee.ucla.edu} \\
  \medskip
  }

\begin{document}

\maketitle

\begin{abstract}
Post-training quantization is an effective method for reducing the serving cost of large language models, where the standard approach is to use a round-to-nearest quantization level scheme. However, this often introduces large errors due to outliers in the weights. 
Proposed mitigation mechanisms include applying adaptive rounding, random rotation transformations or committing to a post-training target using calibration data. Unfortunately, this reliance on calibration data can be severely limiting in some real-world scenarios as such data may be unavailable or subject to privacy regulations.  In this paper, we propose algorithms to optimize transformations and adaptive rounding without access to {\em any} calibration data. The optimization is achieved by designing a suitable proxy function for the quantization loss without calibration data. To maintain inference efficiency, we perform structured matrix transformations for single matrices. For paired weights that interact directly in the computation graph, we use dual matrix transformations and adaptive rounding methods. We conduct experiments on Gemma 2 models, and observe consistent improvement over the baselines. For Gemma 2 9B quantization, our method improves the average benchmark score from 61.9 to 62.4 for 4-bit quantization and from 52.0 to 60.6 for 3-bit quantization, while adding less than 3\% of computation overhead. Furthermore, our method achieves performance comparable to the commonly used GPTQ method, which requires calibration data. 

\end{abstract}

\section{Introduction}

Large language models \citep{brown2020language, chowdhery2022palm, thoppilan2022lamda, touvron2023llama} contain billions or trillions of parameters, which require up to terabytes of memory.
For example, storing a single parameter at standard 16-bit precision (e.g., bfloat16) requires 2 bytes. A model with 100 billion parameters would thus require at least 200 GB of memory just to hold its weights, while a 1 trillion parameter model would require 2 TB. This massive model size far exceeds the on-chip high bandwidth memory (HBM) available on a single accelerator (e.g., GPU), and is complicated by the fact that each forward pass requires reading the entire set of model weights from memory \citep{davies2025efficient}. %
The time spent on loading these parameters into memory often exceeds the time spent on tensor computations, causing memory to be the bottleneck of LLM inference.%

Quantizing weight matrices is a common strategy to reduce the memory loading time, which improves inference efficiency. For example, instead of using 2 bytes  for each parameter, quantization can represent the parameter using just 1 byte (8 bits) or even as few as 4 bits.
The standard method for quantizing weight matrices in LLMs is \emph{uniform} quantization. In uniform quantization with $N$ bits, the weights are divided into blocks (channels)  $\vec{w} = (w_1, \ldots, w_d)$, and then the range of these weights is computed to determine the quantization scale $s = (w_{\rm max} - w_{\rm \min}) / (2^N- 1)$, where $w_{\max} = \max_i {w_i}$ and $ w_{\min}= \min_i {w_i}$. Then, each quantized value is computed as
\begin{align}
\label{eq:uniform_quant}
    \hat{w_i} = s \cdot \left \lceil{\frac{w_i - w_{\rm \min}}{s}}\right\rfloor + w_{\rm \min},
\end{align}
where $\left \lceil x\right\rfloor$ denotes the nearest integer to $x$. The error of uniform quantization over a set of weights indexed by $i$ is upper bounded by
\begin{equation}
\label{eq:q_error_intro}
 | w_i - \hat{w}_i | \leq \frac{ w_{\max}- w_{\min}}{2(2^{N} - 1)} \leq \frac{ \|w \|_\infty}{2^{N} - 1}.
\end{equation}

Despite being widely adopted, uniform quantization suffers from the following drawbacks: 
\begin{inparaenum}[(1)]
\item The error of uniform quantization is determined by the quantization scale, which is heavily affected by the outlier weights within a block. %
\item Each weight matrix in the transformer network is independently quantized. This does not take advantage of the computational structure in the transformer block, particularly in the attention modules.
\end{inparaenum}
which in contrast, aims to quantize an {\em already-trained} model with little to no retraining. PTQ techniques can be differentiated along two broad axes. The first axis is the complexity of the quantization scheme, which ranges from the simplest scalar uniform quantization \citep{frantar2023gptqaccurateposttrainingquantization} to more complex methods like scalar look-up table (LUT) quantization \citep{Squeeze-LLM-ICML-2024}, vector uniform quantization \citep{tseng2024quip}, and vector LUT quantization \citep{tseng2024qtip}. The second axis pertains to whether calibration data is needed at all.

\emph{Goals of this paper.} We focus on calibration data-free scalar uniform quantization. We choose uniform quantization for its high efficiency and broad support on modern hardware accelerators, though our techniques can extend to other quantization formats such as  floating-point quantization \citep{abecassis2025pretraining,mishra2025mxfp8recipes}. Calibration data-free (or {\em calibration-free}, for short) quantization is useful in several scenarios. First, in some applications, representative data for calibration is unavailable
and even when data exists, its use may be prohibited due to privacy and security concerns. Models operating on sensitive information, such as protected health information in medical applications or biometric data for identification, cannot easily use this data for quantization.
Secondly, relying on a static calibration set introduces a significant vulnerability to domain shift. A model may need to be deployed for multiple or unknown downstream tasks, and specifically quantizing for each task may be prohibitive. A model quantized based on a specific dataset can see a degradation in performance when the real-world data it encounters evolves or differs from the calibration set \citep{tang2023easyquant, williams2024impact}.
These motivations underscore the necessity of robust quantization techniques that are {\em data-independent}, thereby ensuring broader applicability, enhanced privacy, and robustness to domain shifts, in downstream applications.

The central questions in scalar PTQ for LLMs revolve around two main challenges:
\begin{asparaenum}[\bfseries (a)]
    \item \textbf{Handling outliers:} How do we prevent a few large magnitude values from dominating the quantization range and destroying precision for all other inliers?
    \item \textbf{Rounding:} How to map high-precision values to their low-precision counterparts?
\end{asparaenum}

\section{Related work}

We refer readers to Appendix~\ref{app:related} for a more comprehensive overview of quantization methods. In this section, we highlight the most relevant works on calibration-free scalar uniform quantization.

\textbf{Handling outliers}: If one applies naive uniform scalar quantization (\Cref{eq:uniform_quant}), which uses the global minimum and maximum of a weight matrix (or sub-matrix), then the error of each parameter would be $O(s)$, which scales linearly in the difference between maximum and minimum values of parameters. However, in large language models this value can be quite large~\citep{chee2024quip2bitquantizationlarge}. 
To overcome this, \citep{dettmers2023case, li2025icquant} proposed to use a small number of bits to quantize inliers and a higher  number of bits for outliers.
A more popular approach is to  multiply the matrix with a random matrix prior to quantization \citep{adepu2024framequantflexiblelowbitquantization, ashkboos2024quarot}. 
A random matrix ensures most values are of the same range and reduces the difference between the minimum and the maximum. This is similar to the phenomenon observed in other quantization works such as gradient compression \citep{Suresh-icml17, vargaftik2022eden} and JL transform \citep{Vempala-RandProj-05, kane2014sparser}.  

\textbf{Rounding techniques:} To the best of our knowledge, all prior calibration-free methods (e.g., \citet{dettmers2023case}) use round-to-the-nearest mapping, where each parameter is mapped to the nearest quantized value. However, we note that there are several results that propose better rounding methods if one assumes that calibration data is available, e.g., \citep{frantar2023gptqaccurateposttrainingquantization, PVtuning-NeurIPS-2024}.

Separately, there are other works that propose calibration-free quantization by using synthetic or distilled data \citep{cai2020zeroq, xu2020generative, sharma2021generalized}. We refer readers to the recent survey of \citet{kim2025zero} for further motivation and techniques in other areas of deep learning. %
However, the works surveyed therein do not focus on LLMs; calibration-free quantization methods for LLMs are less explored, and are the focus of our work.

\section{Our contributions}
\label{sec:prelim}
Modern LLMs consist of stacked layers of transformer blocks, and an embedding layer. Each transformer block includes an attention block and a feedforward-network (FF) block, parameterized by the corresponding weight matrices, and scaling vectors in layer norms. The embedding layer is parameterized by an embedding matrix. Most of the computation in these layers is the matrix-vector multiplication $y = W x$. In this paper, we focus on weight-only quantization that quantizes weight matrices to preserve the result of the output multiplication operations. With this background in mind, we propose a calibration-free PTQ framework that addresses the above mentioned challenges through four primary contributions:

\textbf{Handling outliers via proxy-loss minimization}  A central challenge in calibration-free quantization is the inability to measure the impact of quantization error on downstream task performance. We first establish that the Frobenius norm of the quantization error, $\| W - \widehat{W}\|_F $ (where $W$ is the original matrix and  
$\hat{W}$ is the quantized matrix), serves as a good proxy for final task accuracy. Since the quantization operator, $\widehat{\cdot}$, makes $\| W - \widehat{W}\|_F$ non-differentiable, we bound $\| W - \widehat{W}\|_F$ with surrogate losses to smooth the optimization landscape. %
Combining the above observations, we reframe the problem of mitigating outliers as an optimization problem with the goal of minimizing designed surrogate losses. 
We focus on methods that apply affine transforms to the weight matrices to mitigate outlier effects. Therefore the question becomes: can we learn an affine transformation that minimizes this Frobenius norm quantization error?

\textbf{Structured transformations for single matrices} 
For layers that operate independently, such as feed-forward layers, following earlier works of~\citet{adepu2024framequantflexiblelowbitquantization, ashkboos2024quarot}, we propose to apply an efficient transformation $M$ to the matrix $W$ before quantization and then apply the inverse of the transformation during inference.  The effective layer parameter is thus  $M^{-1} \widehat{M W}$, where as before, $\widehat{\cdot }$ denotes the quantization operation.  However, unlike prior works that use fixed or randomized transformations, we learn a structured transformation. This structured transformation is obtained by minimizing a proxy loss for the quantization error. We find that block-diagonal matrices perform best in our experiments, in contrast to the Walsh-Hadamard transform used in earlier works.

Structured transformations can be applied to all weight matrices, including the attention block, the feedforward block, and the embedding layer. We will discuss this technique in detail in \cref{sec:single}, and how one can choose $\tx$ such that the $\tx^{-1}$ can be applied efficiently.

\textbf{Transformations for coupled matrices.} 
For matrices that are coupled in the computation graph (i.e., they are applied sequentially {\em without} an intermediate non-linearity), we propose learning an arbitrary matrix $M$. This incurs no additional inference overhead, as the transformation can be absorbed into the coupled weight matrices. If $W_1$ and $W_2$ are coupled matrices, we propose to find $M$ such that  the product of the quantized transformed matrices $\widehat{W_1 M^{-1}} \widehat{M W_2}$ closely approximates the original product $ W_1 W_2$. 

To illustrate the usage of this technique, consider attention score computation between locations $i, j$ below:
\[
Attn_{i,j} ={\rm Softmax}\Paren{\frac{X_j^T  \cdot W_q \cdot W_k \cdot x_{i}}{\sqrt{D}}} \odot \Paren{W_o \cdot W_v \cdot X_{i} }.
\]
Note that the attention computation remains unchanged if the product of certain pairs of matrices remain the same, such as the $(W_v, W_o)$ pair and the $(W_q,W_k)$ pair. Note that for any pair of matrices $W_1, W_2$, and any invertible matrix $\tx$, we have 
\[
W_1 W_2 = (W_1 \tx^{-1}) (\tx W_2).
\]
Hence no additional online operation is needed since the output of the operation is already preserved. 
While this technique is incompatible with rotary positional embeddings 
due to non-commutativity, 
recent architectures are moving towards using ``NoPE'', or eliminating positional embeddings
altogether \citep{kazemnejad2023impact},
given evidence that transformers learn to represent position information
from scratch \citep{haviv2022transformer}.
For example, Llama 4's attention layers alternate between RoPE and NoPE~\citep{MetaLlama4Blog2025}.
However, since many architectures still use RoPE, we focus
on applying our paired quantization technique to the output-value projection
pair $(W_v, W_o)$ and leave applications to the query-key pair, $(W_q,W_k)$, as future work.

\textbf{Better adaptive rounding techniques for coupled matrices.} For coupled matrices, we also propose a new alternating adaptive rounding technique that accounts for the matrix product structure. Intuitively, if a value in the first matrix is rounded down, our method attempts to compensate by rounding up corresponding values in the second matrix that interact with it during multiplication. Similar to our previous contribution, this can be applied to $W_o$ and $W_v$ matrices in all transformer architectures, and for the query-key pair, $(W_q,W_k)$, in some architectures.

By using a combination of these methods, we observe consistent improvement over other calibration-free baselines on Gemma 2 models. For Gemma 2 9B quantization, our method improves the average benchmark score from 61.9 to 62.4 for 4-bit quantization and from 52.0 to 60.6 for 3-bit quantization on standard benchmarks, while adding less than 3\% of computation overhead.

The rest of the paper is organized as follows. In Section~\ref{sec:single}, we propose structured transformations for single matrices, and in Section~\ref{sec:coupled}, we propose transformations for coupled matrices. In Section~\ref{sec:joint}, we propose the adaptive rounding technique for coupled matrices. Finally, in Section~\ref{sec:experiments}, we provide ablation studies and results on the Gemma 2 family of models.

\section{Removing outliers through learned linear transformations}

Our quantization approach is motivated by the fact that random rotations remove outliers \citep{adepu2024framequantflexiblelowbitquantization,tseng2024quip,ashkboos2024quarot}, which can reduce quantization error when applied to the weight matrix before uniform quantization. To further remove outliers beyond random rotations, we propose to learn weight-dependent linear transformations. 

More precisely, for a weight matrix of shape $d_1 \times d_2$,  we learn a weight-dependent invertible transformation matrix $M: d_1 \times d_1$, and perform uniform quantization on $M W$ to yield $\widehat{MW}$. During inference, we apply the inverse transformation $M^{-1}$ to a single input feature vector $Y$ of shape $1 \times d_1$, to get $X M^{-1}$, yielding quantized layer output of%
$$
    \hat{Y} = X M^{-1} \widehat{MW}.
$$%
Compared to the desired output $Y = X W$, the output error introduced by the quantization step is%
$$
    \hat{Y} - Y = X M^{-1} \widehat{MW} - X W = X (M^{-1} \widehat{MW} - W).
$$%
If no weight quantization is performed, i.e., $\widehat{\cdot}$ is the identity operation, then $M^{-1} \widehat{MW} - W = M^{-1} MW - W = \mathbf{0}$, preserving the output. When quantization is performed, we want to learn $M$ such that the expected error on the outputs $\EE_Y[\|\hat{Y} - Y\|^2_F]$ is minimized. With a calibration set, we could approximate the expected error with an empirical average. However, this method cannot be applied in the calibration-free setting.
Our approach for learning $M$ relies on the following inequality, which follows from the Cauchy–Schwarz inequality:%
\begin{equation}  
\begin{aligned}
    \EE[\|\hat{Y} - Y \|_F^2] & = \EE[\|X (M^{-1} \widehat{MW} - W)\|_F^2]
    \le \EE[ \|X\|_F^2 \|M^{-1} \widehat{MW} - W\|_F^2] \\
    & = \|M^{-1} \widehat{MW} - W\|_F^2 \EE[\|X\|_F^2]. \label{lem:norm}
\end{aligned} 
\end{equation}
 Assuming that the change of the expectation on the input norm $\EE[\|X\|_F^2]$ due to quantization is small, we minimize the expected error on the outputs by using $\|M^{-1} \widehat{MW} - W\|_F$ as a proxy. Note that this is the same as the $\ell_2$ loss on the vectorized version of $M^{-1} \widehat{MW} - W$. In this paper, we also refer to this proxy loss as the $\ell_2$ loss.

\begin{SCfigure}
    \centering
    \includegraphics[trim={0.3cm 0 0 2cm},clip,width=0.4\linewidth,page=5]{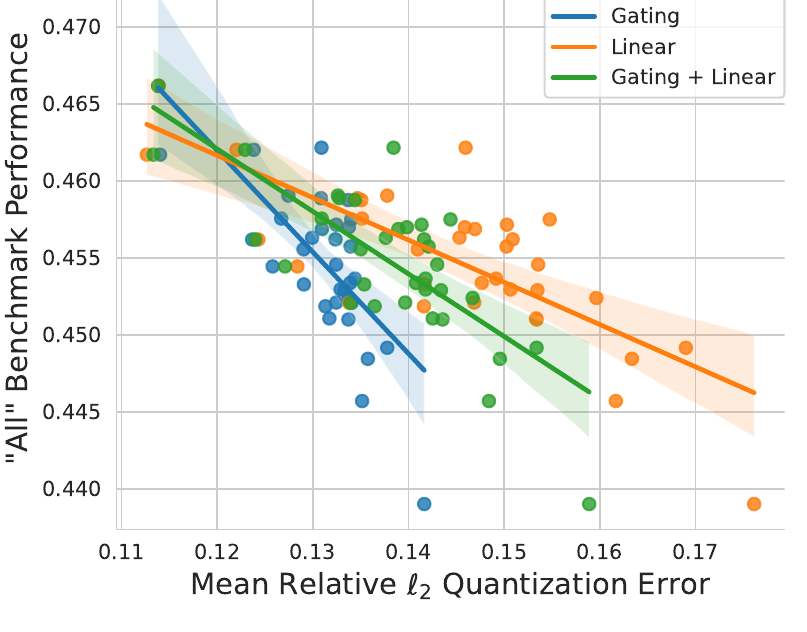}
    \vspace*{-2em}
    \caption{Downstream performance as a function of quantization error of the feedforward  weights. All models were Gemma 2 2B pretrained models whose gating and linear weights were quantized. \emph{Gating + Linear} corresponds to computing the mean quantization error over both sets of weights. We obtain Spearman's rank correlation coefficient of -0.640 (p=4.6e-5) and -0.679 (p=10.0e-6) for the gating and linear layers, respectively.}
\label{fig:all_ff_sweep_intrinsic_vs_extrinsic}
\end{SCfigure}

We perform an empirical study on the feed-forward (FF) block and observe strong Spearman's rank correlation between the proposed $\ell_2$ proxy loss and the scores on downstream evaluations (see \cref{app:l2_downstream_correlation} for details on this analysis). The results are presented in \cref{fig:all_ff_sweep_intrinsic_vs_extrinsic}. The above finding suggests that the Frobenius norm of the reconstruction error on weight matrices is a reasonable proxy loss to use for calibration-free quantization. %
A different challenge we face with this transformation-based approach is the extra computation required due to applying $M^{-1}$ online. In \cref{sec:single} and \cref{sec:coupled}, we discuss how this computation can be mitigated for single and coupled matrices.

\subsection{Learned structured transformation for a single matrix} \label{sec:single}

To reduce the extra inference-time computation, we consider learning structured matrices that support fast matrix-vector multiplication. More specifically, we use block diagonal matrices. A block diagonal matrix of dimension $d$ and block size $k$ can be written as
$$
M = \begin{pmatrix}
    M_1 & & \\
    & \ddots & \\
    & & M_{d/k}
\end{pmatrix}
$$
where all $M_i$'s are $k\times k$ matrices and the neglected entries in $M$ are all zeros. The number of free parameters in a block diagonal matrix is $d \cdot k$. We note that the block diagonal matrices can be multiplied efficiently. During inference, the cost for multiplying such a matrix with a $d$-dimension vector will be $
    C_{d, k} = O(d \cdot k).$

\paragraph{Optimization of $M$.} $\|M^{-1} \widehat{MW} - W\|_F^2$ is not differentiable due to the quantization operator. Hence, for optimizing $M$, we approximate $\|M^{-1} \widehat{MW} - W\|_F^2$ by the expected reconstruction error under the stochastic quantization algorithm. In stochastic quantization, instead of rounding $\frac{w_i-w_{\min}}{s}$ to the nearest integer as in \cref{eq:uniform_quant}, we round to the nearest two integers, each with probability proportional to their distance from the current value. Let $\Delta = \widehat{MW} - MW$.
With stochastic rounding, entries of $\Delta$ are independent of each other with mean zero and variance at most $R^2/(4(2^{N} - 1)^2)$ where $R$ is the range of their corresponding group of weights.
\begin{align}
\EE \left[\|M^{-1} \widehat{MW} - W \|^2_F \right]& = \EE \left[\| M^{-1}(\widehat{MW} - M W)  \|^2_F \right] \nonumber \\
& = \EE \left[\sum^{d_1}_{i=1} \sum^{d_2}_{k=1} \left( \sum^{d_1}_{j=1} (M^{-1})_{i,j} \Delta_{j,k}  \right)^2 \right] \nonumber \\
& =  \sum^{d_1}_{i=1} \sum^{d_2}_{k=1} \sum^{d_1}_{j=1} ((M^{-1})_{i,j})^2 \EE \left[\Delta_{j,k}^2 \right] \nonumber \\
& \leq  \frac{d_2}{(2^{N}-1)^2} \sum^{d_1}_{i=1} \sum^{d_1}_{j=1} ((M^{-1})_{i,j})^2 \|(MW)_j\|_\infty^2. \label{eq:temp_three}
\end{align}
Hence, we initialize $M$ as a block diagonal matrix with each block being a random rotation matrix, and perform gradient descent with \cref{eq:temp_three}  as the loss. Even though we do not perform stochastic quantization in our experiments, the minimizer of \cref{eq:temp_three} yields small quantization error.
In \cref{tab:diag_block_size_first}, we present the average relative $\ell_2$ loss we get with different diagonal block sizes on the Gemma 2 2B model and compare it against uniform quantization and transformed quantization using a random structured matrix. As we can see, we are able to get lower average $\ell_2$ loss compared to both the \textsc{uniform} and \textsc{random}* settings, with a small additional penalty in terms of FLOPs, with diagonal block size $32$. The error is also significantly reduced as the diagonal block size increases. Furthermore, block diagonal matrices can be applied efficiently on GPUs. In \cref{tab:diag_block_size_first}, we also present the extra FLOPs and wall-clock time needed in a GPU implementation compared to the original computation in the down projection layer. We note that applying a block diagonal matrix with diagonal block size 128 requires 1.4\% extra FLOPs and 3.3\% extra wall clock time relative to the original computation. This is the setting we use in our experiments.

\begin{table}[]
    \caption{The effect of block diagonal size on the down projection layer of Gemma 2 2B (input dimension: 9216, output dimension: 2304). \random{} denotes applying a block diagonal matrix with block size 1024, where each block is a random Hadamard matrix. \emph{Extra FLOPs (\%)} and \emph{Extra wall-clock time (\%)} respectively indicate the extra compute and time required compared to the original matmul operation.}
    \label{tab:diag_block_size_first}
    \centering
    \begin{tabular}{ccccccc}
    \hline Method & \uniform & \random&  \multicolumn{4}{c}{Learned block diagonal}\\
    \cmidrule(lr){4-7} 
      Diagonal block size & -  & 1024 &  32 & 64 & 128 & 256 \\
      \hline
      Avg. relative $\ell_2$ loss &0.176 & 0.155 &   0.113 &0.103 & 0.094 & 0.085 \\
      Extra FLOPs (\%) & 0 & 0.11  & 0.35 & 0.69  & 1.39 & 2.78\\
      Extra wall-clock time (\%) & 0 & - & 2.4 & 2.9  & 3.3 & 4.9\\
      \hline
    \end{tabular}

\end{table}

\subsection{Learned paired quantization for coupled matrices} \label{sec:coupled}

As discussed in \cref{sec:prelim}, in the attention computation there are pairs of weight matrices that would not affect the attention output as long as their product is preserved. For these cases, we can couple the transformations of the two matrices so that no online operation is needed.  Let $W_1: d_1 \times d_2$ and $W_2: d_2 \times d_3$ be the weight matrices of two consecutive linear layers, and $\tx$ be an invertible matrix; then we have for any input to the first layer $X \in 1 \times d_1$, 
\[
X W_1 \tx \tx^{-1} W_2  = X W_1 W_2.
\]

Hence, motivated by \cref{lem:norm}, we aim to learn $\tx: d_2 \times d_2$ such that applying quantization on $W_1\tx$ and $\tx^{-1} W_2$ results in a small $\ell_2$ error on matrix products, described below.
\begin{align}
    \min_{\tx, \;\; \widehat{\cdot}} \;\;  \;\;  \| \widehat{W_1 \tx} \widehat{\tx^{-1} W_2} - W_1 W_2  \|_F.
    \label{eq:pqe}
\end{align}
We refer to this quantity as the \emph{paired quantization error} (\pqe). Another design choice in minimizing PQE is the rounding mechanism, which exploits a joint design. In this section, we focus on learning the transformation matrix $\tx$ with the aim of removing outliers, and discuss joint design of the rounding mechanism in \cref{sec:joint}.

Similar to the Frobenius norm in \cref{sec:single}, PQE is also not differentiable.  Hence, we propose to minimize a surrogate loss. Let $U = W_1 \tx$ and $V = \tx^{-1} W_2$. Let the following be channel-wise maximum absolute values of $U$ and $V$, respectively:
\begin{align*}
  m_u(i) = \max_j |U_{i, j}|, \qquad  \qquad m_v(j) = \max_i |V_{i, j}|
\end{align*}
\begin{align*}
 \| \widehat{W_1 \tx} -  W_1 \tx  \|_F +  \| \widehat{\tx^{-1} W_2} - \tx^{-1} W_2 \|_F
 & \leq \| \tx W_1\|_\infty +   \|\tx^{-1} W_2 \|_\infty \\
 & = \|U\|_\infty +   \|V \|_\infty \\
 & \leq \frac{1}{t} \log \left(\sum_{i=1}^{d_1} e^{t m_u(i)} +
       \sum_{j=1}^{d_3} e^{t m_v(j)} \right) \\
       & \triangleq \logsumexp(U, V),
\end{align*}
where the first inequality follows from \cref{eq:q_error_intro} and the second inequality follows from the standard observation that maximum is smaller than log-sum-exp.
Note that $\logsumexp(U, V)$ is a $1$-Lipschitz and $t$-smooth function. We minimize $\logsumexp(U, V)$ over all invertible matrices $M$. Note that since we do not need to do any online rotations during inference, structured matrices are not necessary. Details on various optimization algorithms, regularization parameters, and analysis on optimization dynamics can be found in \cref{app:detailed_study_paired}.

\section{Joint quantization of weight matrices}
\label{sec:joint}

In \cref{sec:coupled}, we described how one can learn a transformation $\tx$ such that the \pqe with individual rounding is minimized. %
In this section, we modify the quantization operator itself (\,$\widehat{\cdot}$\,), to additionally minimize the \pqe defined in \cref{eq:pqe}.

The algorithm takes as inputs two weight matrices and a base rounding map ${\rm Q}$, which permits fast inference-time dequantization. For example, \textit{${\rm Q}$} can be uniform quantization or uniform quantization after applying a transformation learned from the previous step. The algorithm starts by quantizing $W_1$ independently. Then it proceeds iteratively, updating each quantized matrix to compensate for the quantization error in the other matrix.  More specifically, given a quantized version $\widehat{W}_1$ of $W_1$, instead of quantizing $W_2$, it instead quantizes:
\[
W_2' =  \widehat{W_1}^{\dagger} W_1 W_2,
\]
where $\widehat{W_1}^{\dagger}$ is the pseudoinverse of $ \widehat{W_1}$, with the goal of having 
\[
    W_1 W_2' = W_1 \widehat{W_1}^{\dagger} W_1 W_2 \approx  W_1 W_2.
\]
Then we update the quantized version of $W_2$ as
$\widehat{W_2} = {\rm Q}(W_2') = {\rm Q}( \widehat{W_1}^{\dagger} W_1 W_2)$.

\begin{algorithm}[ht]
\caption{Adaptive rounding for matrix product quantization.}
\label{alg:alternative_optimization}
\begin{algorithmic}[1]
\REQUIRE Matrices to be quantized $W_1: d_1 \times h$ and $W_2: h \times d_2$, number of iterations $I$, base quantization function ${\rm Q}$.
\STATE $\widehat{W_1} \leftarrow {\rm Q}(W_1)$ %
\FOR{$i \leftarrow 1$ \TO $I$}
    \STATE $\widehat{W_2} \leftarrow {\rm Q}( \widehat{W_1}^{\dagger} W_1 W_2)$.
    \STATE $\widehat{W_1} \leftarrow {\rm Q}(  W_1 W_2 \widehat{W_2}^{\dagger})$.
\ENDFOR
\RETURN $\widehat{W_1}, \widehat{W_2}$.
\end{algorithmic}
\end{algorithm}

The above iterative quantization scheme still permits fast inference-time dequantization since both $\widehat{W_1}, \widehat{W_2}$ are obtained by the basic rounding scheme $Q$. Moreover, the iterative procedure can produce even lower $\ell_2$ quantization error on the matrix product by explicitly compensating for the quantization loss in the other matrix.

We provide a toy example to illustrate this. Suppose one wants to quantize $2 \times 2$ matrices:
\begin{align*}
    W = W_1 = W_2 &= \begin{bmatrix} 1 & 0 \\ 0 & 0.6  \end{bmatrix}
\end{align*}
to 1 bit, such that the \pqe of $W_1$ and $W_2$ is minimized.
Uniform row-wise or column-wise quantization of either matrix will produce:
$
    \widehat{W} = \begin{bmatrix}
    1 & 0 \\
    0 & 1
    \end{bmatrix} = I,
$
yielding $\widehat{W_1} \widehat{W_2} - W_1 W_2 = \begin{bmatrix} 0 & 0 \\ 0 & 0.64  \end{bmatrix}$ and a \pqe of $\|\widehat{W_1} \widehat{W_2} - W_1 W_2\|_F = 0.64$.
Alternatively, applying even a partial iteration of \cref{alg:alternative_optimization} yields a reduction in the \pqe to $0.36$. To see this, in our iterative quantization scheme, we first quantize $\widehat{W_1} \leftarrow {\rm Q}(W_1) = I$, according to uniform quantization.
Updating $\widehat{W_2} \leftarrow {\rm Q}(\widehat{W_1}^{\dagger} W_1 W_2) = {\rm Q}(I W_1 W_2) = Q \left( \begin{bmatrix} 1 & 0 \\ 0 & 0.36 \end{bmatrix} \right) = \begin{bmatrix} 1 & 0 \\ 0 & 0 \end{bmatrix} $, yielding a \pqe of $\|\widehat{W_1} \widehat{W_2} - W_1 W_2\|_F =0.36$.

We conduct experiments to validate the gain of learned matrix and adaptive rounding for a Gemma 2 2B model, listing the results in \cref{tab:paired_comparison_l2}. The learned transformation reduces $\ell_2$ error by over 16\% relative to both random rotation and independent rounding. Adaptive rounding further reduces $\ell_2$ error by 21\%.
\begin{table}[t]
    \caption{Average relative $\ell_2$ reconstruction error comparison on $W_v W_o$ with 4 bit per-channel quantization. \textrm{I} stands for independent rounding, and \textrm{A} stands for adaptive rounding. Averages are computed over all Gemma 2 2B layers.}
    \centering
    \begin{tabular}{c|cccc}
      \hline
      VO Quantization method & \uniform\ (\textrm{I}) & \random\ (\textrm{I}) & Learned (\textrm{I}) & Learned (\textrm{A})\\
      \hline
      Average $\ell_2$ loss &  0.182 & 0.179 & 0.149 & \textbf{0.117} \\
      \hline
    \end{tabular}
    \label{tab:paired_comparison_l2}
\end{table}

The time complexity of our iterative algorithm at each step is dominated by computing the Moore-Penrose inverse, which itself is dominated by the singular value decomposition (SVD) of each constituent matrix. The time complexity of the SVD of a $d \times h$ matrix is $O\left(d \cdot h \cdot \min(d, h)\right)$ \citep{golub2013matrix}. Therefore, the time complexity of the whole algorithm on input matrices of $d \times h$ , number of iterations $I$, and with ${\rm Q}$ as the uniform quantization operation is 
$O\left(I \cdot d \cdot h \cdot \min(d, h)\right)$.

\section{Experiments}
\label{sec:experiments}
We perform experiments with pretrained Gemma 2 2B and 9B models \citep{gemmateam2024gemma2improvingopen}.  We consider both 3 and 4 bit quantization with each quantization block being a channel along the contraction dimension (per-channel quantization), as well as subchannel quantization with block size 256 and 128. We perform quantization only on the weights and leave the embedding table and activations unquantized.
For CafeQ quantization, we use paired transformation for the pair of  $W_v$ and $W_o$. For all the remaining matrices, we use block diagonal matrices along the contraction dimension with diagonal size $128$ for Gemma 2 2B and $256$ for Gemma 2 9B. In total, this adds $<3\%$ of additional FLOPs for both models.

We evaluate the model on standard Gemma 2 benchmarks~\citep{gemmateam2024gemma2improvingopen}, as described in \cref{tab:task_descs} in the appendix. We present the results in \cref{tab:gemma_quantization}. As shown, CafeQ quantization achieves superior performance compared to the uniform quantization baselines with no rotation or random Hadamard rotations. The improvement is consistent for both the 4-bit case where the quantization loss is smaller and the 3-bit case where the quantization loss is large. We also observe consistent improvement across different quantization block sizes, and find that improvements on individual tasks are also consistent with overall performance (Appendix~\ref{app:per_task}).

\begin{table}[h]
\caption{Average benchmark scores for Gemma 2 2B and 9B quantization. Number of quantization bits and subchannel block sizes are listed along the columns. All evaluations of CafeQ method were conducted on models where the V and O matrices were reverse-transformed prior to inference (even though this was not strictly necessary). $^{*}$ indicates runs (2B 4-bit quantization) where the VO matrices were not transformed back before quantized models were evaluated. Applying the reverse transformation for those runs resulted in similar, but slightly worse overall performance: 45.3, 45.8, 45.9.}
\label{tab:gemma_quantization}
\centering
\setlength{\tabcolsep}{4pt} %
\begin{tabular}{l ccc ccc ccc ccc}
\toprule
\multirow{4}{*}{Method} & \multicolumn{6}{c}{Gemma 2 2B} & \multicolumn{6}{c}{Gemma 2 9B} \\
\cmidrule(lr){2-7} \cmidrule(lr){8-13}
& \multicolumn{3}{c}{3 bits} & \multicolumn{3}{c}{4 bits} & \multicolumn{3}{c}{3 bits} & \multicolumn{3}{c}{4 bits} \\
\cmidrule(lr){2-4} \cmidrule(lr){5-7} \cmidrule(lr){8-10} \cmidrule(lr){11-13}
& N/A & 256 & 128 & N/A & 256 & 128 & N/A & 256 & 128 & N/A & 256 & 128 \\
\midrule
Unquantized & \multicolumn{6}{c}{48.0} & \multicolumn{6}{c}{63.9} \\
\midrule
\uniform & 23.3 & 33.1 & 34.9 & 41.3 & 44.7 & 45.4 & 28.0 & 50.4 & 58.0 & 58.1 & 61.1 & 62.3  \\
\addlinespace
\random & 26.1 & 34.0 & 35.3 & 43.6 & 44.5 & 45.3 & 40.4 & 52.0 & 59.0 & 60.3 & 61.9 & 62.3\\
\addlinespace
\textbf{\cafeq~(ours)} & \textbf{35.1} & \textbf{38.7} & \textbf{39.1} & \textbf{45.6}$^{*}$ & \textbf{46.0}$^{*}$ & \textbf{46.1}$^{*}$ & \textbf{53.7} & \textbf{60.6} & \textbf{61.7} & \textbf{61.9} & \textbf{62.4} & \textbf{62.4}  \\
\bottomrule
\end{tabular}
\end{table}

Next we present ablation studies we performed on the Gemma 2 2B model. All ablations were done with per-channel 4 bit quantization.
\paragraph{Ablation on each separate weight matrices.}
We study the effect of quantization on each of the components in a transformer block, including FF matrices, QK matrices, and VO matrices. For ablation on each of the component, we keep other components unquantized. The results are presented in \cref{tab:cafeq_ablation}. We see that our method achieves consistent improvement for each individual component in the transformer block. We see that quantization leads to the most extreme score drop on FF blocks, and the least drop on QK blocks. This could be explained by the large parameter count of the FF layers. While QK and VO have similar parameter counts, QK only affects the attention scores, which may have less impact on the overall performance.

\begin{table}
    \centering
    \caption{Overall downstream performance after applying uniform/random rotation/CafeQ 4-bit quantization to each set of model weights for pretrained Gemma 2 2B, leaving other weights unquantized. For CafeQ, we use a diagonal block size of 128.%
    }
    \begin{tabular}{c|ccc}
      \toprule
      Method \textbackslash{} Weights & FF & QK & VO  \\
      \hline
      Unquantized & \multicolumn{3}{c}{48.0} \\ 
      \hline
      Uniform & 43.9 & 47.2 & 45.9 \\
      Random & 45.3  & 47.4 & 46.4  \\ 
      CafeQ & {\bf 46.6} & {\bf 47.6} &  {\bf 46.8} \\ 
      \bottomrule
    \end{tabular}
    \label{tab:cafeq_ablation}
    \end{table}

\paragraph{The effect of diagonal block size.} Next we study how the diagonal block size in the transformation matrix would affect the downstream evaluations. We observe in \cref{tab:diag_block_size} that as the block downstream performance improves overall due to the increasing expressiveness of the optimization space. With a diagonal block size of 32, we are already improving over \random~transformation-based method.

\begin{table}[h]
    \caption{The effect of block diagonal size on average scores on downstream evaluations and comparison to the calibration-based method, GPTQ. We perform per-channel 4-bit quantization on Gemma 2 2B model for all methods.}
    \label{tab:diag_block_size}
    \centering
    \begin{tabular}{cccccccc}
    \hline Method & \textsc{uniform} & \textsc{random}* & GPTQ &  \multicolumn{4}{c}{Learned block diagonal}\\
    \cmidrule(lr){5-8} 
      Diagonal block size & -  & 1024 & - &  32 & 64 & 128 & 256 \\
      \hline
      Avg. score  &41.3 & 43.6 & 43.7 &  44.7 & 45.4 & 45.3 & 46.2\\
      \hline
    \end{tabular}
    \label{tab:block_downstream}
\end{table}

\paragraph{Comparison to calibration-based method.} In \cref{tab:block_downstream}, we also compare our method with GPTQ~\citep{frantar2023gptqaccurateposttrainingquantization}, a canonical calibration-based method. We use the GPTModel codebase~\citep{qubitium2024gptqmodel} and perform quantization with 1024 calibration samples from the C4 dataset~\citep{raffel2020c4} as recommended. We find that CafeQ is able to achieve better average score compared to GPTQ. This suggests that calibration-free learning methods are competitive with methods that require calibration data, for quantizing LLMs. We view CafeQ and GPTQ as two orthogonal techniques, as GPTQ focuses on how to better round the matrix with calibration data. Combining both techniques to achieve better performance than either individually offers a promising research direction.
\section{Conclusion}
We propose CafeQ, a calibration-free LLM quantization method for improving uniform quantization with learned transformation matrices and adaptive rounding. We showed that our method consistently improves over other calibration-free baselines while adding minimal extra online computation. We view studying how to combine our method with other calibration-based quantization methods as interesting future directions.

\newpage

\bibliography{references}

\begin{thebibliography}{61}
\providecommand{\natexlab}[1]{#1}
\providecommand{\url}[1]{\texttt{#1}}
\expandafter\ifx\csname urlstyle\endcsname\relax
  \providecommand{\doi}[1]{doi: #1}\else
  \providecommand{\doi}{doi: \begingroup \urlstyle{rm}\Url}\fi

\bibitem[Abecassis et~al.(2025)Abecassis, Agrusa, Ahn, Alben, Alborghetti, Andersch, Arayandi, Bjorlin, Blakeman, Briones, et~al.]{abecassis2025pretraining}
Felix Abecassis, Anjulie Agrusa, Dong Ahn, Jonah Alben, Stefania Alborghetti, Michael Andersch, Sivakumar Arayandi, Alexis Bjorlin, Aaron Blakeman, Evan Briones, et~al.
\newblock Pretraining large language models with nvfp4.
\newblock \emph{arXiv preprint arXiv:2509.25149}, 2025.

\bibitem[Adepu et~al.(2024)Adepu, Zeng, Zhang, and Singh]{adepu2024framequantflexiblelowbitquantization}
Harshavardhan Adepu, Zhanpeng Zeng, Li~Zhang, and Vikas Singh.
\newblock {FrameQuant}: Flexible low-bit quantization for transformers, 2024.
\newblock URL \url{https://arxiv.org/abs/2403.06082}.

\bibitem[Ashkboos et~al.(2024{\natexlab{a}})Ashkboos, Croci, do~Nascimento, Hoefler, and Hensman]{ashkboos2024slicegpt}
Saleh Ashkboos, Maximilian~L. Croci, Marcelo~Gennari do~Nascimento, Torsten Hoefler, and James Hensman.
\newblock Slice{GPT}: Compress large language models by deleting rows and columns.
\newblock In \emph{The Twelfth International Conference on Learning Representations}, 2024{\natexlab{a}}.
\newblock URL \url{https://openreview.net/forum?id=vXxardq6db}.

\bibitem[Ashkboos et~al.(2024{\natexlab{b}})Ashkboos, Mohtashami, Croci, Li, Cameron, Jaggi, Alistarh, Hoefler, and Hensman]{ashkboos2024quarot}
Saleh Ashkboos, Amirkeivan Mohtashami, Maximilian~L. Croci, Bo~Li, Pashmina Cameron, Martin Jaggi, Dan Alistarh, Torsten Hoefler, and James Hensman.
\newblock {QuaRot}: Outlier-free 4-bit inference in rotated {LLM}s.
\newblock In \emph{The Thirty-eighth Annual Conference on Neural Information Processing Systems}, 2024{\natexlab{b}}.
\newblock URL \url{https://openreview.net/forum?id=dfqsW38v1X}.

\bibitem[Austin et~al.(2021)Austin, Odena, Nye, Bosma, Michalewski, Dohan, Jiang, Cai, Terry, Le, and Sutton]{mbpp}
Jacob Austin, Augustus Odena, Maxwell Nye, Maarten Bosma, Henryk Michalewski, David Dohan, Ellen Jiang, Carrie Cai, Michael Terry, Quoc Le, and Charles Sutton.
\newblock Program synthesis with large language models.
\newblock \emph{arXiv preprint arXiv:2108.07732}, 2021.

\bibitem[Bisk et~al.(2020)Bisk, Zellers, Gao, Choi, et~al.]{bisk2020piqa}
Yonatan Bisk, Rowan Zellers, Jianfeng Gao, Yejin Choi, et~al.
\newblock Piqa: Reasoning about physical commonsense in natural language.
\newblock In \emph{Proceedings of the AAAI conference on artificial intelligence}, volume~34, pp.\  7432--7439, 2020.

\bibitem[Brown et~al.(2020)Brown, Mann, Ryder, Subbiah, Kaplan, Dhariwal, Neelakantan, Shyam, Sastry, Askell, et~al.]{brown2020language}
Tom Brown, Benjamin Mann, Nick Ryder, Melanie Subbiah, Jared~D Kaplan, Prafulla Dhariwal, Arvind Neelakantan, Pranav Shyam, Girish Sastry, Amanda Askell, et~al.
\newblock Language models are few-shot learners.
\newblock \emph{Advances in neural information processing systems}, 33:\penalty0 1877--1901, 2020.

\bibitem[Cai et~al.(2020)Cai, Yao, Dong, Gholami, Mahoney, and Keutzer]{cai2020zeroq}
Yaohui Cai, Zhewei Yao, Zhen Dong, Amir Gholami, Michael~W Mahoney, and Kurt Keutzer.
\newblock Zeroq: A novel zero shot quantization framework.
\newblock In \emph{Proceedings of the IEEE/CVF conference on computer vision and pattern recognition}, pp.\  13169--13178, 2020.

\bibitem[Chee et~al.(2024)Chee, Cai, Kuleshov, and Sa]{chee2024quip2bitquantizationlarge}
Jerry Chee, Yaohui Cai, Volodymyr Kuleshov, and Christopher~De Sa.
\newblock {QuIP}: 2-bit quantization of large language models with guarantees, 2024.
\newblock URL \url{https://arxiv.org/abs/2307.13304}.

\bibitem[Chen et~al.(2021)Chen, Tworek, Jun, Yuan, Ponde~de Oliveira~Pinto, Kaplan, Edwards, Burda, Joseph, Brockman, Ray, Puri, Krueger, Petrov, Khlaaf, Sastry, Mishkin, Chan, Gray, Ryder, Pavlov, Power, Kaiser, Bavarian, Winter, Tillet, Petroski~Such, Cummings, Plappert, Chantzis, Barnes, Herbert-Voss, Hebgen~Guss, Nichol, Paino, Tezak, Tang, Babuschkin, Balaji, Jain, Saunders, Hesse, Carr, Leike, Achiam, Misra, Morikawa, Radford, Knight, Brundage, Murati, Mayer, Welinder, McGrew, Amodei, McCandlish, Sutskever, and Zaremba]{humaneval}
Mark Chen, Jerry Tworek, Heewoo Jun, Qiming Yuan, Henrique Ponde~de Oliveira~Pinto, Jared Kaplan, Harri Edwards, Yuri Burda, Nicholas Joseph, Greg Brockman, Alex Ray, Raul Puri, Gretchen Krueger, Michael Petrov, Heidy Khlaaf, Girish Sastry, Pamela Mishkin, Brooke Chan, Scott Gray, Nick Ryder, Mikhail Pavlov, Alethea Power, Lukasz Kaiser, Mohammad Bavarian, Clemens Winter, Philippe Tillet, Felipe Petroski~Such, Dave Cummings, Matthias Plappert, Fotios Chantzis, Elizabeth Barnes, Ariel Herbert-Voss, William Hebgen~Guss, Alex Nichol, Alex Paino, Nikolas Tezak, Jie Tang, Igor Babuschkin, Suchir Balaji, Shantanu Jain, William Saunders, Christopher Hesse, Andrew~N. Carr, Jan Leike, Josh Achiam, Vedant Misra, Evan Morikawa, Alec Radford, Matthew Knight, Miles Brundage, Mira Murati, Katie Mayer, Peter Welinder, Bob McGrew, Dario Amodei, Sam McCandlish, Ilya Sutskever, and Wojciech Zaremba.
\newblock Evaluating large language models trained on code.
\newblock \emph{arXiv preprint arXiv:2107.03374}, 2021.

\bibitem[Chowdhery et~al.(2022)Chowdhery, Narang, Devlin, Bosma, Mishra, Roberts, Barham, Chung, Sutton, Gehrmann, et~al.]{chowdhery2022palm}
Aakanksha Chowdhery, Sharan Narang, Jacob Devlin, Maarten Bosma, Gaurav Mishra, Adam Roberts, Paul Barham, Hyung~Won Chung, Charles Sutton, Sebastian Gehrmann, et~al.
\newblock Palm: Scaling language modeling with pathways.
\newblock \emph{arXiv preprint arXiv:2204.02311}, 2022.

\bibitem[Clark et~al.(2019)Clark, Lee, Chang, Kwiatkowski, Collins, and Toutanova]{boolq}
Christopher Clark, Kenton Lee, Ming-Wei Chang, Tom Kwiatkowski, Michael Collins, and Kristina Toutanova.
\newblock {B}ool{Q}: Exploring the surprising difficulty of natural yes/no questions.
\newblock In Jill Burstein, Christy Doran, and Thamar Solorio (eds.), \emph{Proceedings of the 2019 Conference of the North {A}merican Chapter of the Association for Computational Linguistics: Human Language Technologies, Volume 1 (Long and Short Papers)}, pp.\  2924--2936, Minneapolis, Minnesota, June 2019. Association for Computational Linguistics.
\newblock \doi{10.18653/v1/N19-1300}.
\newblock URL \url{https://aclanthology.org/N19-1300/}.

\bibitem[Clark et~al.(2018)Clark, Cowhey, Etzioni, Khot, Sabharwal, Schoenick, and Tafjord]{clark-2018}
Peter Clark, Isaac Cowhey, Oren Etzioni, Tushar Khot, Ashish Sabharwal, Carissa Schoenick, and Oyvind Tafjord.
\newblock Think you have solved question answering? try {ARC}, the {AI2} reasoning challenge.
\newblock \emph{arXiv preprint arXiv:1803.05457}, 2018.

\bibitem[Cobbe et~al.(2021)Cobbe, Kosaraju, Bavarian, Chen, Jun, Kaiser, Plappert, Tworek, Hilton, Nakano, et~al.]{cobbe-2021}
Karl Cobbe, Vineet Kosaraju, Mohammad Bavarian, Mark Chen, Heewoo Jun, Lukasz Kaiser, Matthias Plappert, Jerry Tworek, Jacob Hilton, Reiichiro Nakano, et~al.
\newblock Training verifiers to solve math word problems.
\newblock \emph{arXiv preprint arXiv:2110.14168}, 2021.

\bibitem[Davies et~al.(2025)Davies, Crago, Sankaralingam, and Kozyrakis]{davies2025efficient}
Michael Davies, Neal Crago, Karthikeyan Sankaralingam, and Christos Kozyrakis.
\newblock Efficient llm inference: Bandwidth, compute, synchronization, and capacity are all you need.
\newblock \emph{arXiv preprint arXiv:2507.14397}, 2025.

\bibitem[Dettmers \& Zettlemoyer(2023)Dettmers and Zettlemoyer]{dettmers2023case}
Tim Dettmers and Luke Zettlemoyer.
\newblock The case for 4-bit precision: k-bit inference scaling laws.
\newblock In \emph{International Conference on Machine Learning}, pp.\  7750--7774. PMLR, 2023.

\bibitem[Dettmers et~al.(2024)Dettmers, Svirschevski, Egiazarian, Kuznedelev, Frantar, Ashkboos, Borzunov, Hoefler, and Alistarh]{SPQr-ICLR-2024}
Tim Dettmers, Ruslan Svirschevski, Vage Egiazarian, Denis Kuznedelev, Elias Frantar, Saleh Ashkboos, Alexander Borzunov, Torsten Hoefler, and Dan Alistarh.
\newblock Spqr: {A} sparse-quantized representation for near-lossless {LLM} weight compression.
\newblock In \emph{International Conference on Learning Representations, {ICLR}}. OpenReview.net, 2024.

\bibitem[Egiazarian et~al.(2024)Egiazarian, Panferov, Kuznedelev, Frantar, Babenko, and Alistarh]{egiazarian2024extremecompressionlargelanguage}
Vage Egiazarian, Andrei Panferov, Denis Kuznedelev, Elias Frantar, Artem Babenko, and Dan Alistarh.
\newblock Extreme compression of large language models via additive quantization, 2024.
\newblock URL \url{https://arxiv.org/abs/2401.06118}.

\bibitem[Frantar et~al.(2023)Frantar, Ashkboos, Hoefler, and Alistarh]{frantar2023gptqaccurateposttrainingquantization}
Elias Frantar, Saleh Ashkboos, Torsten Hoefler, and Dan Alistarh.
\newblock {GPTQ}: Accurate post-training quantization for generative pre-trained transformers, 2023.
\newblock URL \url{https://arxiv.org/abs/2210.17323}.

\bibitem[Golub \& Van~Loan(2013)Golub and Van~Loan]{golub2013matrix}
Gene~H Golub and Charles~F Van~Loan.
\newblock \emph{Matrix computations}.
\newblock JHU press, 2013.

\bibitem[Haviv et~al.(2022)Haviv, Ram, Press, Izsak, and Levy]{haviv2022transformer}
Adi Haviv, Ori Ram, Ofir Press, Peter Izsak, and Omer Levy.
\newblock Transformer language models without positional encodings still learn positional information.
\newblock \emph{arXiv preprint arXiv:2203.16634}, 2022.

\bibitem[Hendrycks et~al.(2021{\natexlab{a}})Hendrycks, Burns, Basart, Zou, Mazeika, Song, and Steinhardt]{hendrycks-iclr-2021}
Dan Hendrycks, Collin Burns, Steven Basart, Andy Zou, Mantas Mazeika, Dawn Song, and Jacob Steinhardt.
\newblock Measuring massive multitask language understanding.
\newblock In \emph{International Conference on Learning Representations, {ICLR}}, 2021{\natexlab{a}}.

\bibitem[Hendrycks et~al.(2021{\natexlab{b}})Hendrycks, Burns, Kadavath, Arora, Basart, Tang, Song, and Steinhardt]{math}
Dan Hendrycks, Collin Burns, Saurav Kadavath, Akul Arora, Steven Basart, Eric Tang, Dawn Song, and Jacob Steinhardt.
\newblock Measuring mathematical problem solving with the math dataset.
\newblock \emph{arXiv preprint arXiv:2103.03874}, 2021{\natexlab{b}}.

\bibitem[Joshi et~al.(2017)Joshi, Choi, Weld, and Zettlemoyer]{triviaqa}
Mandar Joshi, Eunsol Choi, Daniel Weld, and Luke Zettlemoyer.
\newblock {T}rivia{QA}: A large scale distantly supervised challenge dataset for reading comprehension.
\newblock In Regina Barzilay and Min-Yen Kan (eds.), \emph{Proceedings of the 55th Annual Meeting of the Association for Computational Linguistics (Volume 1: Long Papers)}, pp.\  1601--1611, Vancouver, Canada, July 2017. Association for Computational Linguistics.
\newblock \doi{10.18653/v1/P17-1147}.
\newblock URL \url{https://aclanthology.org/P17-1147/}.

\bibitem[Kane \& Nelson(2014)Kane and Nelson]{kane2014sparser}
Daniel~M Kane and Jelani Nelson.
\newblock Sparser johnson-lindenstrauss transforms.
\newblock \emph{Journal of the ACM (JACM)}, 61\penalty0 (1):\penalty0 1--23, 2014.

\bibitem[Kazemnejad et~al.(2023)Kazemnejad, Padhi, Natesan~Ramamurthy, Das, and Reddy]{kazemnejad2023impact}
Amirhossein Kazemnejad, Inkit Padhi, Karthikeyan Natesan~Ramamurthy, Payel Das, and Siva Reddy.
\newblock The impact of positional encoding on length generalization in transformers.
\newblock \emph{Advances in Neural Information Processing Systems}, 36:\penalty0 24892--24928, 2023.

\bibitem[Kim et~al.(2025{\natexlab{a}})Kim, young Kim, Cho, Lee, Kim, and Jeon]{kim2025boaattentionawareposttrainingquantization}
Junhan Kim, Ho~young Kim, Eulrang Cho, Chungman Lee, Joonyoung Kim, and Yongkweon Jeon.
\newblock {BoA}: Attention-aware post-training quantization without backpropagation, 2025{\natexlab{a}}.
\newblock URL \url{https://arxiv.org/abs/2406.13474}.

\bibitem[Kim et~al.(2025{\natexlab{b}})Kim, Choi, Lee, Cho, and Kang]{kim2025zero}
Minjun Kim, Jaehyeon Choi, Jongkeun Lee, Wonjin Cho, and U~Kang.
\newblock Zero-shot quantization: A comprehensive survey.
\newblock \emph{arXiv preprint arXiv:2505.09188}, 2025{\natexlab{b}}.

\bibitem[Kim et~al.(2024)Kim, Hooper, Gholami, Dong, Li, Shen, Mahoney, and Keutzer]{Squeeze-LLM-ICML-2024}
Sehoon Kim, Coleman Hooper, Amir Gholami, Zhen Dong, Xiuyu Li, Sheng Shen, Michael~W. Mahoney, and Kurt Keutzer.
\newblock Squeezellm: Dense-and-sparse quantization.
\newblock In \emph{International Conference on Machine Learning, {ICML}}. OpenReview.net, 2024.

\bibitem[Kwiatkowski et~al.(2019)Kwiatkowski, Palomaki, Redfield, Collins, Parikh, Alberti, Epstein, Polosukhin, Devlin, Lee, Toutanova, Jones, Kelcey, Chang, Dai, Uszkoreit, Le, and Petrov]{nq}
Tom Kwiatkowski, Jennimaria Palomaki, Olivia Redfield, Michael Collins, Ankur Parikh, Chris Alberti, Danielle Epstein, Illia Polosukhin, Jacob Devlin, Kenton Lee, Kristina Toutanova, Llion Jones, Matthew Kelcey, Ming-Wei Chang, Andrew~M. Dai, Jakob Uszkoreit, Quoc Le, and Slav Petrov.
\newblock Natural questions: A benchmark for question answering research.
\newblock \emph{Transactions of the Association for Computational Linguistics}, 7:\penalty0 452--466, 2019.
\newblock \doi{10.1162/tacl_a_00276}.
\newblock URL \url{https://aclanthology.org/Q19-1026/}.

\bibitem[Li et~al.(2025{\natexlab{a}})Li, Hanna, Fragouli, and Diggavi]{li2025icquant}
Xinlin Li, Osama Hanna, Christina Fragouli, and Suhas Diggavi.
\newblock {ICQuant}: Index coding enables low-bit llm quantization.
\newblock \emph{arXiv preprint arXiv:2505.00850}, 2025{\natexlab{a}}.

\bibitem[Li et~al.(2025{\natexlab{b}})Li, Hanna, Fragouli, and Diggavi]{ICQuant-arxiv-2025}
Xinlin Li, Osama~A. Hanna, Christina Fragouli, and Suhas~N. Diggavi.
\newblock Icquant: Index coding enables low-bit {LLM} quantization.
\newblock \emph{CoRR}, abs/2505.00850, 2025{\natexlab{b}}.
\newblock URL \url{https://doi.org/10.48550/arXiv.2505.00850}.

\bibitem[Liu et~al.(2024)Liu, Zhao, Fedorov, Soran, Choudhary, Krishnamoorthi, Chandra, Tian, and Blankevoort]{liu2024spinquant}
Zechun Liu, Changsheng Zhao, Igor Fedorov, Bilge Soran, Dhruv Choudhary, Raghuraman Krishnamoorthi, Vikas Chandra, Yuandong Tian, and Tijmen Blankevoort.
\newblock {SpinQuant: LLM} quantization with learned rotations, 2024.
\newblock URL \url{https://arxiv.org/abs/2405.16406}.

\bibitem[Ma et~al.(2024)Ma, Li, Zheng, Ling, Xiao, Wang, Wen, Chao, and Ji]{ma2024affinequantaffinetransformationquantization}
Yuexiao Ma, Huixia Li, Xiawu Zheng, Feng Ling, Xuefeng Xiao, Rui Wang, Shilei Wen, Fei Chao, and Rongrong Ji.
\newblock Affinequant: Affine transformation quantization for large language models, 2024.
\newblock URL \url{https://arxiv.org/abs/2403.12544}.

\bibitem[Malinovskii et~al.(2024)Malinovskii, Mazur, Ilin, Kuznedelev, Burlachenko, Yi, Alistarh, and Richt{\'{a}}rik]{PVtuning-NeurIPS-2024}
Vladimir Malinovskii, Denis Mazur, Ivan Ilin, Denis Kuznedelev, Konstantin Burlachenko, Kai Yi, Dan Alistarh, and Peter Richt{\'{a}}rik.
\newblock Pv-tuning: Beyond straight-through estimation for extreme {LLM} compression.
\newblock In \emph{Advances in Neural Information Processing Systems ({NeurIPS})}, 2024.

\bibitem[{Meta}(2025)]{MetaLlama4Blog2025}
{Meta}.
\newblock The llama 4 herd: The beginning of a new era of natively multimodal ai innovation.
\newblock \url{https://ai.meta.com/blog/llama-4-multimodal-intelligence/}, April 2025.
\newblock Accessed: 2025-09-24.

\bibitem[Mishra et~al.(2025)Mishra, Stosic, Layton, and Micikevicius]{mishra2025mxfp8recipes}
Asit Mishra, Dusan Stosic, Simon Layton, and Paulius Micikevicius.
\newblock Recipes for pre-training llms with mxfp8.
\newblock \emph{arXiv preprint arXiv:2506.08027}, 2025.

\bibitem[ModelCloud.ai(2024)]{qubitium2024gptqmodel}
ModelCloud.ai.
\newblock Gpt-qmodel.
\newblock \url{https://github.com/modelcloud/gptqmodel}, 2024.
\newblock Contact: qubitium@modelcloud.ai.

\bibitem[Raffel et~al.(2020)Raffel, Shazeer, Roberts, Lee, Narang, Matena, Zhou, Li, and Liu]{raffel2020c4}
Colin Raffel, Noam Shazeer, Adam Roberts, Katherine Lee, Sharan Narang, Michael Matena, Yanqi Zhou, Wei Li, and Peter~J. Liu.
\newblock Exploring the limits of transfer learning with a unified text-to-text transformer.
\newblock \emph{Journal of Machine Learning Research}, 21\penalty0 (140):\penalty0 1--67, 2020.
\newblock URL \url{http://jmlr.org/papers/v21/20-074.html}.

\bibitem[Sakaguchi et~al.(2021)Sakaguchi, Bras, Bhagavatula, and Choi]{winogrande}
Keisuke Sakaguchi, Ronan~Le Bras, Chandra Bhagavatula, and Yejin Choi.
\newblock Winogrande: An adversarial winograd schema challenge at scale.
\newblock \emph{Communications of the ACM}, 64\penalty0 (9):\penalty0 99--106, 2021.

\bibitem[Sap et~al.(2019)Sap, Rashkin, Chen, Le~Bras, and Choi]{siqa}
Maarten Sap, Hannah Rashkin, Derek Chen, Ronan Le~Bras, and Yejin Choi.
\newblock Social {IQ}a: Commonsense reasoning about social interactions.
\newblock In Kentaro Inui, Jing Jiang, Vincent Ng, and Xiaojun Wan (eds.), \emph{Proceedings of the 2019 Conference on Empirical Methods in Natural Language Processing and the 9th International Joint Conference on Natural Language Processing (EMNLP-IJCNLP)}, pp.\  4463--4473, Hong Kong, China, November 2019. Association for Computational Linguistics.
\newblock \doi{10.18653/v1/D19-1454}.
\newblock URL \url{https://aclanthology.org/D19-1454/}.

\bibitem[Shao et~al.(2024)Shao, Chen, Zhang, Xu, Zhao, Li, Zhang, Gao, Qiao, and Luo]{shao2024omniquant}
Wenqi Shao, Mengzhao Chen, Zhaoyang Zhang, Peng Xu, Lirui Zhao, Zhiqian Li, Kaipeng Zhang, Peng Gao, Yu~Qiao, and Ping Luo.
\newblock Omniquant: Omnidirectionally calibrated quantization for large language models.
\newblock In \emph{The Twelfth International Conference on Learning Representations}, 2024.
\newblock URL \url{https://openreview.net/forum?id=8Wuvhh0LYW}.

\bibitem[Sharma et~al.(2021)Sharma, Abraham, and Rajendiran]{sharma2021generalized}
Prasen~Kumar Sharma, Arun Abraham, and Vikram~Nelvoy Rajendiran.
\newblock A generalized zero-shot quantization of deep convolutional neural networks via learned weights statistics.
\newblock \emph{IEEE Transactions on Multimedia}, 25:\penalty0 953--965, 2021.

\bibitem[Sun et~al.(2025)Sun, Liu, Bai, Bao, Zhao, Li, Hu, Yu, Hou, Yuan, Jiang, Liu, and Yao]{sun2025flatquantflatnessmattersllm}
Yuxuan Sun, Ruikang Liu, Haoli Bai, Han Bao, Kang Zhao, Yuening Li, Jiaxin Hu, Xianzhi Yu, Lu~Hou, Chun Yuan, Xin Jiang, Wulong Liu, and Jun Yao.
\newblock Flatquant: Flatness matters for llm quantization, 2025.
\newblock URL \url{https://arxiv.org/abs/2410.09426}.

\bibitem[Suresh et~al.(2017)Suresh, Yu, Kumar, and McMahan]{Suresh-icml17}
Ananda~Theertha Suresh, Felix~X. Yu, Sanjiv Kumar, and H.~Brendan McMahan.
\newblock Distributed mean estimation with limited communication.
\newblock In \emph{Proceedings of the 34th International Conference on Machine Learning,{ICML}}, volume~70 of \emph{Proceedings of Machine Learning Research}, pp.\  3329--3337. {PMLR}, 2017.

\bibitem[Suzgun et~al.(2023)Suzgun, Scales, Sch{\"a}rli, Gehrmann, Tay, Chung, Chowdhery, Le, Chi, Zhou, and Wei]{bbh}
Mirac Suzgun, Nathan Scales, Nathanael Sch{\"a}rli, Sebastian Gehrmann, Yi~Tay, Hyung~Won Chung, Aakanksha Chowdhery, Quoc Le, Ed~Chi, Denny Zhou, and Jason Wei.
\newblock Challenging {BIG}-bench tasks and whether chain-of-thought can solve them.
\newblock In Anna Rogers, Jordan Boyd-Graber, and Naoaki Okazaki (eds.), \emph{Findings of the Association for Computational Linguistics: ACL 2023}, pp.\  13003--13051, Toronto, Canada, July 2023. Association for Computational Linguistics.
\newblock \doi{10.18653/v1/2023.findings-acl.824}.
\newblock URL \url{https://aclanthology.org/2023.findings-acl.824/}.

\bibitem[Tang et~al.(2023)Tang, Sun, Wu, Liu, Zhu, and Kang]{tang2023easyquant}
Hanlin Tang, Yifu Sun, Decheng Wu, Kai Liu, Jianchen Zhu, and Zhanhui Kang.
\newblock Easyquant: An efficient data-free quantization algorithm for llms.
\newblock In \emph{Proceedings of the 2023 Conference on Empirical Methods in Natural Language Processing}, pp.\  9119--9128, 2023.

\bibitem[Team(2024)]{gemmateam2024gemma2improvingopen}
Gemma Team.
\newblock Gemma 2: Improving open language models at a practical size, 2024.
\newblock URL \url{https://arxiv.org/abs/2408.00118}.

\bibitem[Thoppilan et~al.(2022)Thoppilan, De~Freitas, Hall, Shazeer, Kulshreshtha, Cheng, Jin, Bos, Baker, Du, et~al.]{thoppilan2022lamda}
Romal Thoppilan, Daniel De~Freitas, Jamie Hall, Noam Shazeer, Apoorv Kulshreshtha, Heng-Tze Cheng, Alicia Jin, Taylor Bos, Leslie Baker, Yu~Du, et~al.
\newblock Lamda: Language models for dialog applications.
\newblock \emph{arXiv preprint arXiv:2201.08239}, 2022.

\bibitem[Touvron et~al.(2023)Touvron, Lavril, Izacard, Martinet, Lachaux, Lacroix, Rozi{\`e}re, Goyal, Hambro, Azhar, et~al.]{touvron2023llama}
Hugo Touvron, Thibaut Lavril, Gautier Izacard, Xavier Martinet, Marie-Anne Lachaux, Timoth{\'e}e Lacroix, Baptiste Rozi{\`e}re, Naman Goyal, Eric Hambro, Faisal Azhar, et~al.
\newblock {LLaMA}: Open and efficient foundation language models.
\newblock \emph{arXiv preprint arXiv:2302.13971}, 2023.

\bibitem[Tseng et~al.(2024{\natexlab{a}})Tseng, Chee, Sun, Kuleshov, and Sa]{tseng2024quip}
Albert Tseng, Jerry Chee, Qingyao Sun, Volodymyr Kuleshov, and Christopher~De Sa.
\newblock Qu{IP}\${\textbackslash}\#\$: Even better {LLM} quantization with hadamard incoherence and lattice codebooks.
\newblock In \emph{Forty-first International Conference on Machine Learning}, 2024{\natexlab{a}}.
\newblock URL \url{https://openreview.net/forum?id=9BrydUVcoe}.

\bibitem[Tseng et~al.(2024{\natexlab{b}})Tseng, Chee, Sun, Kuleshov, and Sa]{tseng2024quipbetterllmquantization}
Albert Tseng, Jerry Chee, Qingyao Sun, Volodymyr Kuleshov, and Christopher~De Sa.
\newblock {QuIP\#}: Even better {LLM} quantization with hadamard incoherence and lattice codebooks, 2024{\natexlab{b}}.
\newblock URL \url{https://arxiv.org/abs/2402.04396}.

\bibitem[Tseng et~al.(2024{\natexlab{c}})Tseng, Sun, Hou, and Sa]{tseng2024qtip}
Albert Tseng, Qingyao Sun, David Hou, and Christopher~De Sa.
\newblock {QTIP}: Quantization with trellises and incoherence processing.
\newblock In \emph{The Thirty-eighth Annual Conference on Neural Information Processing Systems}, 2024{\natexlab{c}}.
\newblock URL \url{https://openreview.net/forum?id=7sdkLVuYCU}.

\bibitem[van Breugel et~al.(2025)van Breugel, Bondarenko, Whatmough, and Nagel]{vanbreugel2025fptquantfunctionpreservingtransformsllm}
Boris van Breugel, Yelysei Bondarenko, Paul Whatmough, and Markus Nagel.
\newblock Fptquant: Function-preserving transforms for llm quantization, 2025.
\newblock URL \url{https://arxiv.org/abs/2506.04985}.

\bibitem[Vargaftik et~al.(2022)Vargaftik, Basat, Portnoy, Mendelson, Itzhak, and Mitzenmacher]{vargaftik2022eden}
Shay Vargaftik, Ran~Ben Basat, Amit Portnoy, Gal Mendelson, Yaniv~Ben Itzhak, and Michael Mitzenmacher.
\newblock Eden: Communication-efficient and robust distributed mean estimation for federated learning.
\newblock In \emph{International Conference on Machine Learning}, pp.\  21984--22014. PMLR, 2022.

\bibitem[Vempala(2005)]{Vempala-RandProj-05}
S.~Vempala.
\newblock \emph{The random projection method}.
\newblock American Mathematical Society, 2005.

\bibitem[Williams \& Aletras(2024)Williams and Aletras]{williams2024impact}
Miles Williams and Nikolaos Aletras.
\newblock On the impact of calibration data in post-training quantization and pruning.
\newblock In \emph{Proceedings of the 62nd Annual Meeting of the Association for Computational Linguistics (Volume 1: Long Papers)}, pp.\  10100--10118, 2024.

\bibitem[Xiao et~al.(2023)Xiao, Lin, Seznec, Wu, Demouth, and Han]{xiao2023smoothquant}
Guangxuan Xiao, Ji~Lin, Mickael Seznec, Hao Wu, Julien Demouth, and Song Han.
\newblock {S}mooth{Q}uant: Accurate and efficient post-training quantization for large language models.
\newblock In \emph{Proceedings of the 40th International Conference on Machine Learning}, 2023.

\bibitem[Xu et~al.(2020)Xu, Li, Zhuang, Liu, Cao, Liang, and Tan]{xu2020generative}
Shoukai Xu, Haokun Li, Bohan Zhuang, Jing Liu, Jiezhang Cao, Chuangrun Liang, and Mingkui Tan.
\newblock Generative low-bitwidth data free quantization.
\newblock In \emph{European conference on computer vision}, pp.\  1--17. Springer, 2020.

\bibitem[Zellers et~al.(2019)Zellers, Holtzman, Bisk, Farhadi, and Choi]{hellaswag}
Rowan Zellers, Ari Holtzman, Yonatan Bisk, Ali Farhadi, and Yejin Choi.
\newblock {H}ella{S}wag: Can a machine really finish your sentence?
\newblock In Anna Korhonen, David Traum, and Llu{\'i}s M{\`a}rquez (eds.), \emph{Proceedings of the 57th Annual Meeting of the Association for Computational Linguistics}, pp.\  4791--4800, Florence, Italy, July 2019. Association for Computational Linguistics.
\newblock \doi{10.18653/v1/P19-1472}.
\newblock URL \url{https://aclanthology.org/P19-1472/}.

\bibitem[Zhong et~al.(2024)Zhong, Cui, Guo, Liang, Lu, Wang, Saied, Chen, and Duan]{zhong-naacl-2024}
Wanjun Zhong, Ruixiang Cui, Yiduo Guo, Yaobo Liang, Shuai Lu, Yanlin Wang, Amin Saied, Weizhu Chen, and Nan Duan.
\newblock {AGIEval}: A human-centric benchmark for evaluating foundation models.
\newblock In \emph{Findings of the Association for Computational Linguistics: NAACL 2024}, pp.\  2299--2314, 2024.

\end{thebibliography}
\bibliographystyle{iclr2026_conference}

\appendix

\section{Related work}
\label{app:related}

The idea of transforming weight matrices to make them easier to quantize has been widely explored in the literature. SmoothQuant~\citep{xiao2023smoothquant} and OmniQuant~\citep{shao2024omniquant} use calibration data to apply activation-dependent scaling on weight matrices.

Applying random rotations to weight matrices has been shown to be an effective way of mitigating outliers prior to quantization. This idea has been explored in recent works to improve the performance of uniform quantization~\citep{adepu2024framequantflexiblelowbitquantization, ashkboos2024quarot, chee2024quip2bitquantizationlarge, liu2024spinquant}. This has been extended to use vector quantizers instead of scalar quantizers \cite{tseng2024qtip,tseng2024quip}. Using non-uniform scalar quantizers has been studied in \citep{PVtuning-NeurIPS-2024}. All these methods use calibration data to design the quantizers. There are methods that do not use rotations, but try to index outliers directly \citep{SPQr-ICLR-2024,Squeeze-LLM-ICML-2024,ICQuant-arxiv-2025}, but these methods require one to index outliers and also require calibration data for good performance. %
To reduce the inference-time computation to rotate the matrices back, two main techniques have been used: (1) Exploiting the computational invariance property~\citep{ashkboos2024slicegpt, ashkboos2024quarot, liu2024spinquant, ma2024affinequantaffinetransformationquantization} in transformer blocks to avoid online rotations for ``paired'', composed, matrices, (2) Rotate the weight matrices using a restricted class of structured matrices \citep{adepu2024framequantflexiblelowbitquantization, tseng2024quipbetterllmquantization, ashkboos2024quarot} such as Hadamard matrices, that admit fast online rotations.

Methods have also been proposed to introduce correlation between the quantization for different weights and channels to further reduce quantization error. For example, \citet{frantar2023gptqaccurateposttrainingquantization, chee2024quip2bitquantizationlarge, kim2025boaattentionawareposttrainingquantization} use a calibration dataset to calculate the Hessian matrix of the output error with respect to each model weight, and use the Hessian matrix to introduce correlations between the quantization function of different weights. While most works quantize each layer independently, \citet{liu2024spinquant,tseng2024quipbetterllmquantization, egiazarian2024extremecompressionlargelanguage} use finetuning to introduce cross-weight dependencies to further reduce the quantization error. 

Similar to our work, FlatQuant~\citep{sun2025flatquantflatnessmattersllm} also considers learned structured matrices for better quantization of non-paired matrices. They use the Kronecker product of two lightweight matrices for faster online reverse transformation. The concurrent work of FPTQuant~\citep{vanbreugel2025fptquantfunctionpreservingtransformsllm} considers learning non-rotation matrices for the transformation matrix on $(W_v, W_o)$ pair. However, all these learning-based methods rely on calibration data for optimizing the transformation.

\section{Evaluation datasets}

\begin{table}[ht]
\centering
\caption{Downstream benchmarks considered in this work. We report the average performance over \emph{All} of these tasks, along with average performance over the \emph{Core}, \emph{Math}, and \emph{Code} subsets. We use top-1 accuracy as the metric for all non-coding tasks, where the generated prefix for sampling tasks must match the reference. For coding tasks, we consider Pass@1 as the metric.}
\label{tab:task_descs}
    \begin{tabular}{llllll}
        \toprule
        Task & Core & Math & Code & Notes & Eval Setting \\ \midrule
        MMLU \citep{hendrycks-iclr-2021} & x & x &  & 5-shot & scoring \\
        ARC (Challenge) \citep{clark-2018} & x & x &  & 0-shot & scoring \\
        GSM8K \citep{cobbe-2021} & x & x &  & 8-shot & sampling \\
        AGIEval (English) \citep{zhong-naacl-2024} & x &  &  & 3-5-shot & sampling \\
        BBH \citep{bbh} & x & x &  & 0-shot & sampling \\
        Winogrande \citep{winogrande} & x & x &  &  & scoring \\
        HellaSwag \citep{hellaswag} & x &  &  & 0-shot & scoring \\
        MATH \citep{math} &  & x &  & 4-shot & sampling \\
        ARC (Easy) \citep{clark-2018} &  &  &  & 0-shot & scoring \\
        PIQA \citep{bisk2020piqa} &  &  &  &  & scoring \\
        SIQA \citep{siqa} &  &  &  &  & scoring \\
        BoolQ \citep{boolq} &  &  &  & 0-shot & scoring \\
        TriviaQA \citep{triviaqa} &  &  &  & 5-shot & sampling \\
        NQ \citep{nq} &  &  &  & 5-shot & sampling \\
        HumanEval \citep{humaneval} &  &  & x &  & sampling \\
        MBPP \citep{mbpp} &  &  & x & 3-shot & sampling \\
        \bottomrule
    \end{tabular}
\end{table}

\cref{tab:task_descs} lists the evaluation tasks considered in this work, along with how they are grouped into task suites. Performance on individual tasks and suites are presented in \cref{app:per_task}. \

\section{Details for feedforward parameter sweep} \label{app:l2_downstream_correlation}

Here we provide sweep details for producing \cref{fig:all_ff_sweep_intrinsic_vs_extrinsic}. In this case we considered 4-bit quantization of the gating and linear weights of the feedforward network. We report model performance averaged over \emph{All} downstream tasks, and swept over various quantization hyperparameters in order to produce a range of models with varying intrinsic and extrinsic performance. We considered models without any quantization, vanilla uniform quantization, random rotation/Hadamard transformation prior to quantization, in addition to a range of models with learned block Hadamard transformations. For all learning runs, we set the number of iterations to 20,000, and swept over the following parameters for learning transformations of the FFW:

\begin{itemize}
  \item Learning rate: $\{0.1, 0.5, 1.0\}$
  \item Block diagonal size: $\{2, 4, 8, 16, 64, 256\}$
  \item Number of Hadamard matrices: $\{0, 1, 2\}$
\end{itemize}

Each point in \cref{fig:all_ff_sweep_intrinsic_vs_extrinsic} corresponds to the relative quantization error with respect to a particular subset of FFW (gating, linear, or the mean of the two) against the downstream performance for one of these models.

\section{Paired quantization hyperparameter details}  \label{app:detailed_study_paired}

\subsection{Alternative choices for the pseudo-loss}
Initially we considered other pseudo-loss choices for learning paired transformations. Using the notation described in \cref{sec:coupled}, we considered (1) the mean of squared channel-wise maxima:

\begin{align*}
  \frac{\sum_{i=1}^{m} (M^u_i)^2}{m} + \frac{\sum_{j=1}^{n} (M^v_j)^2}{n}
  \frac{\sum_{i=1}^{d_1} m_u(i)^2}{d_1} + \frac{\sum_{j=1}^{d_3} m_v(j)^2}{d_3}
\end{align*}

and (2) the weighted mean of squared channel-wise maxima:

\begin{align*}
  \frac{\|V'\|_F \sum_{i=1}^{d_1} m_u(i)^2}{d_1} + \frac{\|U'\|_F \sum_{j=1}^{d_3} m_v(j)^2}{d_3}
\end{align*}

This latter objective is motivated by the intuition that quantization error in one weight matrix has the potential to be magnified if the norm of its paired weight is large. Thus, quantization errors should be penalized accordingly. This objective explicitly acknowledges the fact that $U'$ and $V'$ directly interact in the computation graph. We sweep over the choice of this pseudo-loss in the following subsection.

\subsection{Hyperparameter tuning} \label{app:hyperparameter_tuning}
\begin{figure}
  \centering
  \includegraphics[width=0.7\linewidth]{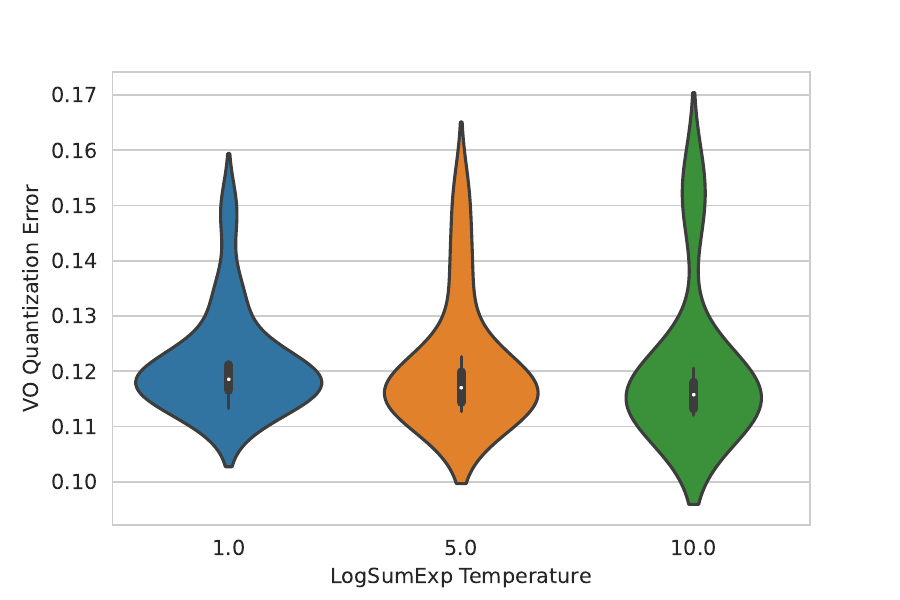}
  \caption{Distribution over the VO product relative \pqe as a function of the temperature on the LogSumExp pseudo-loss, optimized with Adam. Each violin encompasses a sweep over learning rate in $10^{\{-4, -3, -2, -1\}}$, and orthogonal regularization weight in $\{0, 0.1, 1\}$ for the given value of $t$.  Note that we exclude runs with learning rate of 1.0, as these runs diverged (\cref{fig:tune_lr}).}
  \label{fig:tune_ploss_temp}
\end{figure}

We explored several hyperparameters for learning paired VO transformations. In this section, we present sweeps over these hyperparameters, noting which ones the \pqe happened to be sensitive to. We fix number of iterations to 100,000 for all runs, and consider Cayley SGD (with $\beta=0.1$ momentum) as well unconstrained optimization with Adam ($\beta=0.1$), learning $\tx$ to transform V/O for all layers in a Gemma2-2B-PT model. We compute the mean \pqe across all layers, assuming uniform channel-wise 4-bit quantization for $\widehat{\cdot}$. Note that we report the relative \pqe in this section -- relative to $\|W_v^T W_o\|_F$. This ensures that the average \pqe is not unduly influenced by the magnitude of $W_v^T W_o$ for any given layer.

We varied:

 \begin{itemize}
 \item base learning rate $lr \in 10^{\{-4, -3, -2, -1, 0\}}$ 
 \item orthonormal regularization weight $\lambda_{orth} \in \{0, 0.01, 0.1, 1\}$
 \item LogSumExp temperature $t \in \{1, 5, 10\}$
 \item Pseudo-loss $\in \{$LogSumExp, Sum of Squares, Weighted Sum of Squares$\}$
\end{itemize}

\paragraph{LogSumExp temperature} \pqe as a function of $t$, the temperature of the LogSumExp pseudo-loss, is shown in \cref{fig:tune_ploss_temp}. While the mean is relatively unchanged across settings of $t$, the variance of the ultimate \pqe increases with $t$. We chose $t=5$ for LogSumExp pseudo-loss experiments, as \pqe was relatively insensitive to this hyperparameter.

\begin{figure*}
  \centering
  \begin{subfigure}[t]{0.9\textwidth}
  \includegraphics[page=4,width=\textwidth]{figures/cayley_lr_learning_curves.pdf}
  \subcaption{Cayley SGD.}
  \end{subfigure}
  \begin{subfigure}[t]{0.9\textwidth}
  \includegraphics[page=1,width=\textwidth]{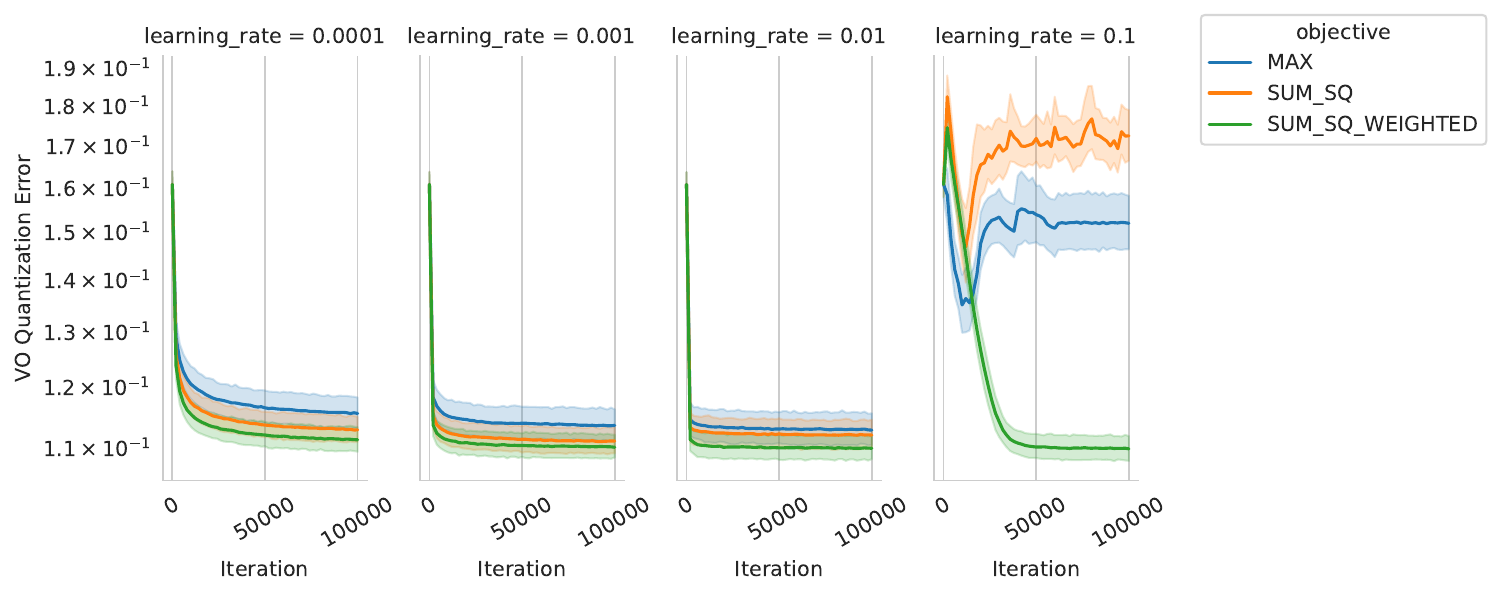}
  \subcaption[t]{Adam; $\lambda_{orth}=0$.}
  \end{subfigure}
  \begin{subfigure}[t]{0.9\textwidth}
  \includegraphics[page=4,width=\textwidth]{figures/adam_lr_by_orth_learning_curves.pdf}
  \subcaption{Adam; $\lambda_{orth}=1$.}
  \end{subfigure}
  \caption{Learning curves for Cayley SGD (top) and Adam with various orthonormal regularization weights (center and bottom), with each pseudo-loss a separate line. Learning rate is varied along columns. \pqe is on the y-axis and iteration count on the x-axis. Each line corresponds to the mean \pqe across all model layers for a given pseudo-loss, with the 95\% bootstrap confidence interval indicated by the shaded region.}
\label{fig:tune_lr}
\end{figure*}

\paragraph{Learning rate \& Pseudo-loss} While $t$ had little effect on \pqe, both the base learning rate and choice of pseudo-loss affected the stability of the learning curve. \cref{fig:tune_lr} displays learning curves for Cayley SGD and Adam as a function of learning rate and pseudo-loss.

\pqe tends to converge quickly with Adam assuming that the learning rate is small enough (less than $0.1$). This is true irrespective of the choice of pseudo-loss. While Cayley SGD also smoothly reduces \pqe, it benefits from a higher learning rate for \pqe to converge in the allotted iterations.

\paragraph{Constraints on $\tx$} We observe that $\tx$ need only be invertible to ensure that the transformation is computationally invariant. Prior work learns $\tx$ such that $\tx$ is strictly a rotation matrix \citep{liu2024spinquant}. In this work, we varied the degree to which $\tx$ is orthonormal
by minimizing a pseudo-loss directly with Adam without strictly enforcing orthonormality. We experimented with adding a regularization term to the loss to encourage $\tx$ to not stray too far from orthonormality:
\[
 \frac{\| \tx \tx^T - I \|_F}{\sqrt{d}}.
\]

We controlled the weight of this term with $\lambda_{orth}$

\begin{table}
\centering
\caption{Relative mean \pqe across all layers after 100K iterations, as a function of objective, optimization method, and $\lambda_{orth}$. The average minimum/maximum eigenvalue across all layers for the learned $\tx$ are given as well.}
\label{tab:min_error_over_orth}
\begin{tabular}{llrrrr}
\toprule
Objective & Opt & $\lambda_{\text{orth}}$ & \text{PQE} & $\min{\sigma}$ & $\max{\sigma}$ \\ \midrule
SumSqWted & Adam & 0.00 & 0.110 & 338.848 & 894.407 \\
LogSumExp & Adam & 0.10 & 0.118 & 0.978 & 1.062 \\
LogSumExp & Adam & 0.01 & 0.118 & 0.845 & 2.175 \\
LogSumExp & Adam & 0.00 & 0.120 & 0.876 & 2.508 \\
LogSumExp & Adam & 1.00 & 0.120 & 0.981 & 1.049 \\
SumSqWted & Cayley & 0.00 & 0.134 & 0.992 & 1.000 \\
LogSumExp & Cayley & 0.00 & 0.142 & 0.997 & 1.001 \\
SumSq & Cayley & 0.00 & 0.152 & 0.999 & 1.001 \\
SumSqWted & Adam & 0.01 & 0.166 & 2.503 & 912.959 \\
SumSqWted & Adam & 0.10 & 0.185 & 1.915 & 905.954 \\
SumSqWted & Adam & 1.00 & 0.200 & 1.460 & 730.075 \\
SumSq & Adam & 0.10 & 0.255 & 0.867 & 165.328 \\
SumSq & Adam & 0.01 & 0.261 & 0.886 & 222.902 \\
SumSq & Adam & 1.00 & 0.462 & 0.869 & 138.006 \\
SumSq & Adam & 0.00 & 0.639 & 0.551 & 971.175 \\
\bottomrule
\end{tabular}
\end{table}

\paragraph{Importance of learning an orthonormal $\tx$} Although prior work has learned $\tx$ such that $\tx$ is a rotation, in practice, $\tx$ need only be invertible to ensure computational invariance. To determine how important it is that $\tx$ be orthonormal, we select the minimum \pqe run per objective, optimization method, and orthonormal regularization weight, while sweeping over base learning rate, as well as $t$ for the LogSumExp runs.

\cref{tab:min_error_over_orth} displays the relative \pqe for each of these runs along with the minimum and maximum eigenvalue, averaged across all layers, for the learned $\tx$. The lowest \pqe transformations are learned through unconstrained, Adam, optimization without any weight on the orthonormal regularization term, with a mean relative \pqe of 0.110 vs. 0.118 for the next best run. Note also that the average minimum and maximum eigenvalue for these transformations are quite large, suggesting the learned transformations are far from rotations.It is important to also note that learning strict rotation transformations via Cayley SGD yields mean \pqe which is at least 21.7\% higher than the best unconstrained learning run.

Note that \pqe is an intrinsic measure of quality, and one must also evaluate these quantized models downstream to note any degradation. In terms of stability with respect to $\lambda_{\text{orth}}$, learning transformations that are close to rotation, and achieving low \pqe, unconstrained optimization with the LogSumExp pseudo-loss yields a better solution than weighted sum of squares or learning $\tx$ using natural gradient descent. With that in mind, we selected unconstrained optimization with LogSumExp pseudo-loss unless mentioned otherwise.

\section{Downstream performance on individual tasks}
\label{app:per_task}

\begin{table}[ht]
\centering
\large
\caption{Individual task performance along with task suite averages for various Gemma 2 2B quantized models. The best-performing quantized model on each task/suite is noted in bold. Performance of the original Gemma 2 2B is given in the \emph{None} column.}
\label{tab:per_task_perf}
\begin{tabular}{l|c|cccc}
\toprule
Benchmark & None & GPTQ & Uniform & Random & CafeQ \\
\midrule
MMLU & 51.9 & 48.2 & 43.9 & 47.5 & \textbf{50.1} \\
ARC-C & 50.3 & 47.9 & 46.3 & \textbf{49.7} & 49.0 \\
GSM8K & 21.2 & 12.4 & 9.5 & 13.3 & \textbf{18.6} \\
AGIEval & 31.7 & 26.9 & 26.5 & 27.2 & \textbf{28.9} \\
BBH & 42.6 & 37.8 & 35.0 & 34.3 & \textbf{38.8} \\
Winogrande & 68.6 & \textbf{67.8} & 67.4 & 67.5 & 66.9 \\
HellaSwag & 73.9 & 71.3 & 69.5 & 70.5 & \textbf{71.9} \\ \midrule
MATH & 16.0 & 9.0 & 8.5 & 10.0 & \textbf{13.2} \\
ARC-e & 80.7 & 78.2 & 76.3 & 78.7 & \textbf{78.9} \\
PIQA & 78.5 & 77.0 & 76.9 & 77.6 & \textbf{78.0} \\
SIQA & 51.6 & 50.8 & 49.5 & 51.2 & \textbf{51.4} \\
Boolq & 73.0 & \textbf{73.4} & 63.6 & 72.4 & 67.1 \\
TriviaQA & 60.3 & 52.3 & 48.5 & 49.4 & \textbf{55.2} \\
NQ & 17.3 & 13.4 & 11.5 & 12.8 & \textbf{15.2} \\
HumanEval & 19.5 & 14.6 & 9.8 & \textbf{15.2} & 14.6 \\
MBPP & 30.4 & 18.0 & 17.6 & 18.8 & \textbf{26.0} \\ \midrule
Average (All) & 48.0 & 43.7 & 41.3 & 43.5 & \textbf{45.2} \\
Average (Core) & 48.6 & 44.6 & 42.6 & 44.3 & \textbf{46.3} \\
Average (Math) & 41.8 & 37.2 & 35.1 & 37.0 & \textbf{39.4} \\
Average (Code) & 25.0 & 16.3 & 13.7 & 17.0 & \textbf{20.3} \\
\bottomrule
\end{tabular}
\end{table}

\cref{tab:per_task_perf} displays the performance of CafeQ against competing quantization methods on each individual task. CafeQ outperforms all other quantization methods on each task suite: all, core, math, and coding tasks. It also achieves the highest quantized performance on 12 out of 16 of the individual tasks, suggesting that overall performance is not driven by outperformance on a small subset of tasks.

\end{document}